# Training Efficient CNNS: Tweaking the Nuts and Bolts of Neural Networks for Lighter, Faster and Robust Models

Sabeesh Ethiraj
UpGrad Education Pvt Limited,Mumbai
sabeesh90@yahoo.co.uk

Bharath Kumar Bolla
Salesforce, Hyderabad
bolla111@gmail.com

*Abstract*—Deep Learning has revolutionized the fields of computer vision, natural language understanding, speech recognition, information retrieval and more. Many techniques have evolved over the past decade that made models lighter, faster, and robust with better generalization. However, many deep learning practitioners persist with pre-trained models and architectures trained mostly on standard datasets such as Imagenet, MS-COCO, IMDB-Wiki Dataset, and Kinetics-700 and are either hesitant or unaware of redesigning the architecture from scratch that will lead to better performance. This scenario leads to inefficient models that are not suitable on various devices such as mobile, edge, and fog. In addition, these conventional training methods are of concern as they consume a lot of computing power. In this paper, we revisit various SOTA techniques that deal with architecture efficiency (Global Average Pooling, depth-wise convolutions & squeeze and excitation, Blurpool), learning rate (Cyclical Learning Rate), data augmentation (Mixup, Cutout), label manipulation (label smoothing), weight space manipulation (stochastic weight averaging), and optimizer (sharpness aware minimization). We demonstrate how an efficient deep convolution network can be built in a phased manner by sequentially reducing the number of training parameters and using the techniques mentioned above. We achieved a SOTA accuracy of 99.2% on MNIST data with just 1500 parameters and an accuracy of 86.01% with just over 140K parameters on the CIFAR-10 dataset.

## I. INTRODUCTION

For the past decade, deep learning with neural networks has been the most popular way for training new machine learning models. The ImageNet competition in 2012 is widely credited with its rise to fame. A deep convolutional network AlexNet [1] fared 41% better than the next best entry that year. As a result of this ground-breaking study, a race to build deeper networks with an ever-increasing number of parameters and complexity ensued. Several model architectures, including VGGNet [2], Inception [3], ResNet [4], and others, have consistently beaten prior marks in ImageNet contests over the years while also expanding their footprint (model size, latency, etc.). Progressive improvements on benchmarks like image classification, text classification, and so on have been correlated with an increase in network complexity, the number of parameters, amount of training resources required to train the network, prediction latency, and so on since deep learning research has been focused on improving the state-of-the-art.

While these models can perform effectively on the tasks for which they were trained, they may not be suitable for immediate deployment in the real world. Building efficient deep learning models is imperative given the hardware and practical constraints. Efficient models can be broadly classified based on two categories; 1. Inference efficiency and 2. Training efficiency. Inference efficiency deals with how many parameters the model has, how large the model is, how much RAM is consumed during inference, how long the inference delay is, etc. On the other hand, training efficiency deals with how long it takes for a model to train, how many devices there are, if the model is memory efficient, etc.

However, we may not need to optimize for any given scenario for both types of efficiencies. In this paper, we have revisited techniques that pertain to increased inference efficiency. Our work will help the industrial and academic community to get introduced to the latest techniques and help them implement the same in their field of work

The objective of this paper is as follows:

1) To make models more inference efficient by improving architectural efficiency.

2) To make models more robust by data augmentation.

3) To improve model generalizability by tweaking optimization algorithms, manipulating labels, and altering weight space.

## II. RELATED WORK

Much work has been done in the preceding years on making models more and more efficient as the impact of deep learning models has become increasingly evident lately.

### A. Architectural Efficiency

The role of depth-wise convolutions in improving architectural efficiency was first established in the Xception network [5]. Depth-wise, separable convolutions were later incorporated in the MobileNet [6] architecture to build lighter models. Depth-wise separable convolutions decrease the number of training parameters without compromising the dimensionality of the features/channels extracted compared to a conventional 3x3 kernel. While old school CNN architectures use dense layers at the end of convolutional blocks to cater for increased learning ability via an increase in parameters, this leads to flattening of the network, which compromises a neural network's ability to localize the features extracted by the preceding convolutional blocks as reflected in work done by [7]. Similarly, the work done by [8] showed that while the earlier layers capture only low-level features, the higher layers capture task-specific features. There is a need to preserve these features extracted, which were made possible by the concept of Global Average Pooling, wherein the information from the convolution blocks are condensed via averaging and are at the same time linearized before classification. Though feature extraction and preservation is an inherent task of convolutional layers, techniques such as max-pooling result in the loss of shift equivariance due to sub-sampling. To counter this, the concept of anti-aliasing filters was introduced in the pioneering work by [9] that could render a neural network more shift-invariant, subsequently making the model more robust. The deeper layers of a neural network are highly dimensional. However, only specific channels are hypothesized to contain related yet vast amounts of information for a given classification task at a given depth. Hence attention to these specific channels must be given if a deep network is expected to come out with a definitive output. The same was elaborated in work done by [10] wherein the channel outputs were passed through a sequence of dense layers with sigmoid activation to arrive at different scalar weights, which are later applied to the corresponding channels in a '*squeeze and excite*' manner. Similar works on optimizing architecture were carried out wherein custom models were built using layer fine tuning techniques [11], augmentations [12] and custom modification of CNNs [13] that improve efficiency of a model.

*B. Data Augmentation*

Several data augmentation techniques have been innovated over the last decade, resulting in increasing a model's overall performance. State of the art results was obtained on CIFAR-10, CIFAR-100, and SVHN datasets using techniques such as cutouts [14], where random regions of an input image were masked out during training to make the models more accurate, robust, and generalizable. Similar work on augmentation done by [15] introduced the concept of mixup where input images and corresponding target variables were combined in varying proportions, resulting in completely new virtual training data used to improve model robustness.

*C. Optimization and Regularization*

In any neural network, the outcome that determines the accuracy of a model are fundamentally the weights that are learned over a considerable number of epochs. The work done by [16] showed that simply averaging the weights over a constant or a cyclical learning rate schedule resulted in better model performance. 75% of the training is done using the conventional training procedure, after which stochastic weight averaging is done as per the learning rate schedule. Another research done along the lines of optimization was the sharpness awareness minimization by [17]. Instead of just minimizing the training loss, the sharpness of the loss in neighboring loss space was also minimized, preventing the model from reaching localized minima, resulting in more generalizability. The revolutionary work [18] on label smoothing in a multi-class scenario where a fraction modified the target variable revealed that this technique prevented a model's overconfidence, subsequently increasing its robustness.

III. METHODOLOGY

State-of-the-art techniques such as Architectural efficiency enhancement, Weight Space Alterations, Label manipulation, Optimization, and Faster training have been used in the Mosaic ML [19] library experiments.

*A. Dataset*

MNIST and CIFAR 10 datasets have been used to conduct the experiments mentioned. The datasets consist of 50000 training and 10000 test data points.

*B. Parameter Reduction - Depth Wise Separable Convolutions*

Depth-wise convolutions reduce the number of training parameters due to their mathematical efficiency of depth-wise channel separation and point-wise convolutions. A sample representation of depth-wise separable convolutions with a 3x3 filter on a 3-channel image is shown in Fig 1, showing a reduction in the number of parameters.

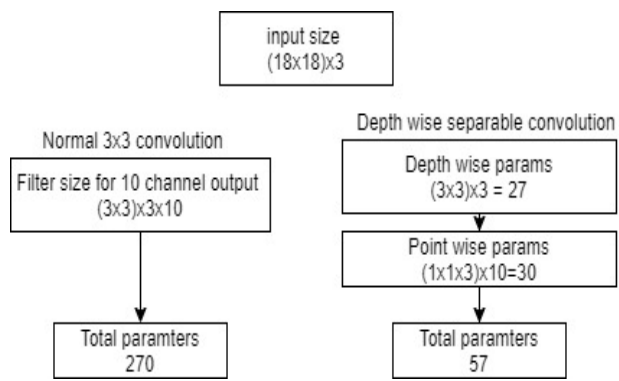

Fig 1. Depth Wise Separable Convolutions

*C. Parameter Reduction - Global Average pooling*

Global average Pooling (GAP) is performed by taking the average of all the neurons/pixels in a channel for all the channels to linearize the output of a convolution.

*D. Channel Attention - Squeeze and Excite*

Squeeze and Excitation (SE) blocks consist of two separate mechanisms which add attention to specific channels from the output of the previous convolutional layer. The initial squeeze mechanism is done by performing a Global Average Pooling to the output channels followed by an excitation mechanism performed by adding MLP layers with a sigmoid activation to generate a linear scalar weight. These scalar weights are applied to the feature maps or channels to generate the output of the SE block. In our experiments, we have added the SE block at the end of every convolutional layer using appropriate '*latent channel*' and '*min channel*' hyperparameters of the Mosaic ML library.

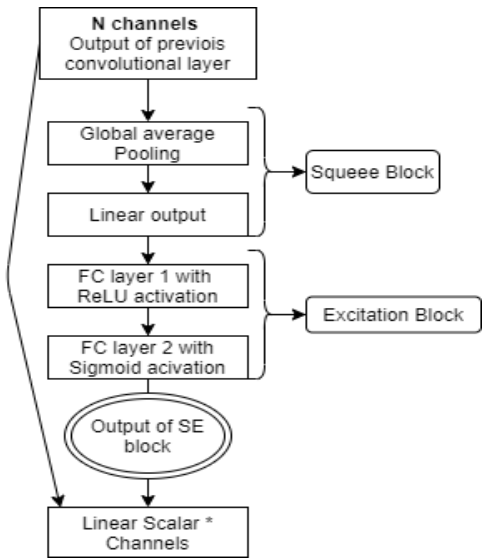

Fig 2. Squeeze and Excitation Blocks

### E. Weight Space Alterations – Stochastic Weight Averaging

Stochastic Weight Averaging (SWA) is a regularization method that works on the principle of cyclical learning rate. There are two models. The first model stores the average weight of the models at the end of each learning rate cycle schedule of the second model, while the second model runs the conventional training algorithm. The final predictions are based on the stochastic weights average stored in the first model. In our experiments, we have implemented SWA after 75% of the training time, after which the weights are averaged after every training epoch

### F. Anti-Aliasing techniques – Blurpool

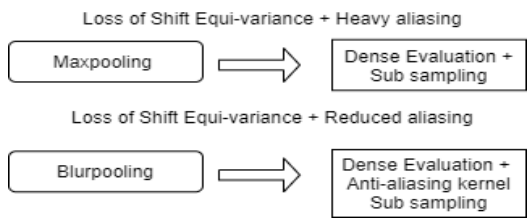

Fig 3. Blurpool

Convolutions and Maxpooling are susceptible to translational or shift variance of the input image as they, in a manner, cause downsampling of the input data resulting in aliasing of the input signals. Blurpool overcomes these deficiencies by introducing anti-aliasing at the time of downsampling. While conventional max-pooling is considered a combination of densely evaluated pooling and subsampling, blur-pooling is done by introducing a blur filter after the densely evaluated pooling. This prevents loss of information resulting in anti-aliasing and resilience to shift-variance. The proposed methodology is described in Figure 3. We have implemented this technique in our experiments both at the convolutional and max-pooling layers

### G. Cutout

The cutout is a regularization or augmentation technique performed by clipping pixels from the input image. This results in better learning as this helps regularize the model. We have implemented this technique by clipping random masks of the input image of dimension 10x10 pixel.

### H. Mixup

Mixup is a regularization technique that works on the principles of combining different input samples along with their target labels to create a new set of virtual training examples. The amount of mixup between the images and the labels is controlled by a hyperparameter $\delta$ which ranges between 0 and 1. The mathematical intuition of this technique taken from the original paper is shown below in Equation 1 where $\hat{x}$ and $\hat{y}$ are new virtual distributions created from existing training images and their corresponding labels.

$$\hat{x} = \delta x_i + (1 - \delta) x_j$$
$$\hat{y} = \delta y_i + (1 - \delta) y_j$$

Equation 1. Mixup Virtual distribution

### I. Sharpness Awareness Minimization

Sharpness Awareness minimization is an optimization algorithm that minimizes both the training loss and the loss sharpness. Instead of minimizing just the loss, which may result in the gradient descent algorithm approaching the local minima, the SAM aims to approach the global minima whose neighborhoods have uniformly low training values in a given loss space. This results in better model generalization. Visualization of the loss landscapes as depicted in the original paper is shown in Figure 4.

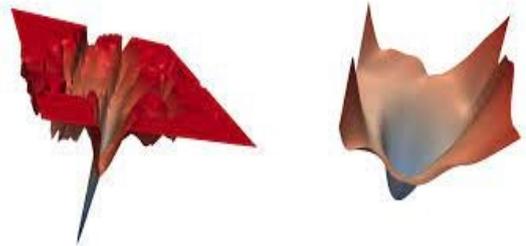

Fig 4. Sharpness Aware minimization with uniformly low neighborhood training loss (Right)

### J. Label smoothing

Label smoothing is a regularization technique that changes the predicted target variable by a small quantity $\alpha$. This prevents the model from overconfidence of a particular predicted target variable. A hyperparameter $\alpha$ controls the smoothening. In our experiments, we run with $\alpha = 0.1$ to study the least possible label smoothing effect

### K. One Cycle Learning Rate

In this technique, the learning rate is gradually increased from a minimum learning rate to a maximum learning rate during the first half of the cycle and then brought down to a value lower than the initial lower learning rate during the second half of the cycle. The higher learning rate towards the middle of the training helps prevent a sharp decrease in loss and subsequent overfitting. In contrast, the reduction in learning rate towards the end of the second half of the cycle helps achieve the local minima of the loss space much more efficiently. The change in learning rate takes place after every batch.

Table 1. MNIST model architectures

| GAP Models | 25 K without depth-wise convolutions | 7K without depth-wise convolutions | 7k with depth-wise convolutions | 5K without depth-wise convolutions | 5K with depth-wise convolutions | 1.5K with depth-wise convolutions |
|---|---|---|---|---|---|---|
| Params | 25,144 | 7,767 | 7,611 | 5,712 | 5,616 | 1,560 |
| Layers | Conv 1 (1,8)/3x3<br>Conv 2 (8,16)/3x3<br>Maxpool1<br>Conv 3 (16,32)/3x3<br>Conv 4 (32,64)/3x3<br>Maxpool2<br>GAP<br>Conv 5 (64,10)/1x1<br>(*Classifier*) | Conv 1 (1,8)/3x3<br>Conv 2 (8,12)/3x3<br>TB Conv (12,8)/1x1<br>Maxpool1<br>Conv 3 (8,12)/3x3<br>Conv 4 (12,16)/3x3<br>TB Conv (16,12)/1x1<br>Maxpool2<br>Conv 5 (12,15)/3x3<br>Conv 6 (15,15)/3x3<br>GAP<br>Conv 7 (15,10)/1x1<br>(*Classifier*) | Conv 1 (1,10)/3x3<br>**Conv 2 DW (10,12)/3x3**<br>TB Conv (12,10)/1x1<br>Maxpool1<br>Conv 3 (10,13)/3x3<br>Conv 4 (13,16)/3x3<br>TB Conv (16,12)/1x1<br>Maxpool2<br>Conv 5 (12,15)/3x3<br>Conv 6 (15,15)/3x3<br>GAP<br>Conv 7 (15,10)/1x1<br>(*Classifier*) | Conv 1 (1,8)/3x3<br>Conv 2 (8,12)/3x3<br>TB Conv (12,8)/1x1<br>Maxpool1<br>Conv 3 (8,12)/3x3<br>Conv 4 (12,16)/3x3<br>TB Conv (16,12)/1x1<br>Maxpool2<br>Conv 5 (12,15)/3x3<br>GAP<br>Conv 6 (15,10)/1x1<br>(*Classifier*) | Conv 1 (1,8)/3x3<br>**Conv 2 DW (8,12)/3x3**<br>TB Conv (12,8)/1x1<br>Maxpool1<br>Conv 3 (8,13)/3x3<br>Conv 4 (13,18)/3x3<br>TB Conv (18,12)/1x1<br>Maxpool2<br>Conv 5 (12,16)/3x3<br>GAP<br>Conv 6 (16,10)/1x1<br>(*Classifier*) | Conv 1 (1,8)/3x3<br>**Conv 2 DW (8,12)/3x3**<br>TB Conv (12,8)/1x1<br>Maxpool1<br>**Conv 3 DW (8,12)/3x3**<br>**Conv 4 DW (12,16)/3x3**<br>TB Conv (16,12)/1x1<br>Maxpool2<br>**Conv 5 DW (12,15)/3x3**<br>GAP<br>Conv 6 (15,10)/1x1<br>(*Classifier*) |
| DW | No | No | Yes | No | Yes | Yes |
| GAP | Yes | Yes | Yes | Yes | Yes | Yes |

(*TB – Transition block / DW – Depth wise separable convolutions/ GAP – Global Average Pooling layer)

### L. MNIST Model Architectures

Four different model architectures were built, sequentially reducing the number of parameters using depth-wise separable convolutions and Global Average Pooling layer; Model with 25K params, 7K params, 5K params, and 1.5K params. The summary of the architectures is shown in Table 1.

### M. CIFAR 10 Model Architectures

Table 2. CIFAR-10 architectures

| GAP Models | 143 K Without DW | 143 K with DW |
|---|---|---|
| Params | 143,208 | 143,396 |
| Layers | Conv 1 (3,16)/3x3<br>Conv 2 (16,16)/3x3<br>Maxpool1<br>Conv 3 (16,32)/3x3<br>Conv 4 (32,32)/3x3<br>Maxpool2<br>Conv 5 (32,64)/3x3<br>Conv 6 (64,64)/3x3<br>Maxpool3<br>Conv 7 (64,120)/3x3<br>GAP<br>Conv 8 (120,10)/1x1<br>(*Classifier*) | Conv 1 (3,16)/3x3<br>Conv 2 (16,16)/3x3<br>Maxpool1<br>Conv 3 DW (16,32)/3x3<br>Conv 4 (32,32)/3x3<br>Maxpool2<br>Conv 5 DW (32,64)/3x3<br>Conv 6 (64,64)/3x3<br>Maxpool3<br>Conv 7 DW (64,128)/3x3<br>Conv 8 DW (128,192)/3x3<br>Conv 9 DW (192,260)/3x3<br>GAP<br>Conv 2 (260,10)/1x1<br>(*Classifier*) |
| DW | No | No |
| GAP | Yes | Yes |

In the case of CIFAR-10 experiments, we have built two models with Global Average Pooling Layers. The summary of the architectures is shown in Table 2.

### N. Experimental Methodologies

Building these architectures is to sequentially reduce the number of parameters using GAP and depth-wise separable convolutions and to arrive at a model with the least number of parameters with a balanced accuracy – parameters trade-off. These models are later used to perform experimental studies using the various model enhancement techniques mentioned above by studying the effect of these using metrics such as Accuracy, inference time and, model size. The inference time of the models is calculated on the test dataset of 10,000 data points using the CPU as the inference engine.

### O. Loss Function

The loss function used here is the cross-entropy loss as this is a multi-class classification scenario

$$CE = -\sum_{i}^{C} t_i \log(s_i)$$

Equation 2. Cross-Entropy Loss

### P. Evaluation Metrics

The evaluation metrics used in this study are Validation Accuracy, Inference time, and Model size.

## IV. RESULTS

### A. The efficiency of Depth Wise Convolutions on Accuracy

From Tables 1,2, and 3, it is evident that depth-wise (DW) convolutions in combination with GAP help reduce the number of training parameters, but do they actually help increase model performance? It has been one of the prime objectives of this research. To test this hypothesis, i.e., the efficiency of the network in terms of accuracy, the number of parameters is increased in our DW CNNS to match the CNNs without DW convolutions. It is observed that the addition of DW CNNs does not compromise model performance in terms of accuracy despite the reduction in the number of parameters; instead, they perform better or equally well as that of a network with 3x3 convolution in 25K, 7K, and 5K models. This is evident when we achieved a **SOTA accuracy** of 98.35% in the architecture with 1.5K parameters which are as good as a model with 7K or 25K parameters. Similar performances were also seen on the CIFAR-10 dataset, where the model performed equally well as that of a 3x3 convolutional network. Hence, it can be inferred that **depth-wise separable convolutions can help achieve higher accuracy with fewer parameters**. Furthermore, there is also a reduction in model size using DW convolutions due to a reduction in the number of parameters, as seen in Table 3

Table 3. MNIST /CIFAR10 - Accuracy, Latency, and Model Size

| MNIST Dataset | | | | |
|---|---|---|---|---|
| Model | DW | Acc | Latency | Size |
| 25K (25,144K) | No | 99.35 | 3.36 s | 107 KB |
| 7K (7,767K) | No | 99.34 | 2.67 s | 43 KB |
| 7K (7,611K) | Yes | 99.46 | 3.35 s | 43 KB |
| 5K (5,712K) | No | 99.33 | 2.76 s | 34 KB |
| 5K (5,616K) | Yes | 99.27 | 2.74 s | 34 KB |
| **1.5K (1,560K)** | **Yes** | **98.35** | **2.52 s** | **19 KB** |
| CIFAR 10 Dataset | | | | |
| 143K (143,208K) | No | 81.57 | 8.51 s | 578 KB |
| 143K (143,396K) | Yes | 79.9 | 8.82 s | 590 KB |

### B. Effect of Depth Wise Convvolutions on Inference Time

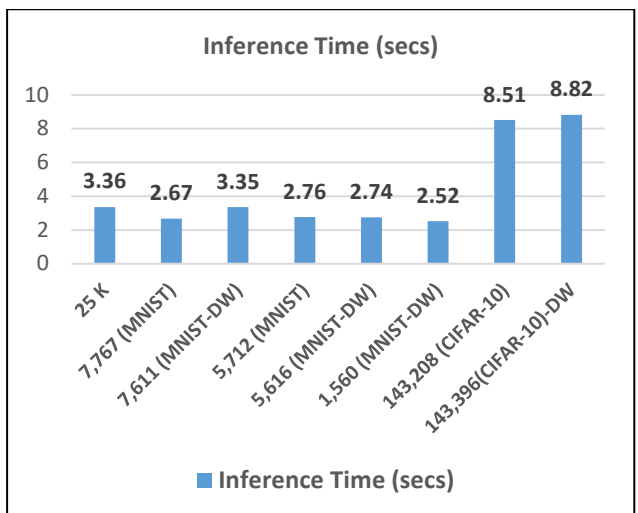

Fig 5. Inference time of Various models - 10000 Test Datapoints

Depth wise convolutions do not have a significant beneficial effect on the inference time of the model. In fact, models with **depth-wise convolutions perform slower than models with 3x3 kernels**. This is evident on both the MNIST and CIFAR-10 datasets, where we see increased latency (Table 3). It is hypothesized that this may be attributed to the two separate convolutional operations (depth-wise separable + point-wise convolution) that consume more mathematical computing time than a single 3x3 convolution. But to achieve faster and efficient models with depth-wise convolutions, there has to be a reduced parameter count and as seen in our experiments, the model with the least number of parameters **(1.5K) using depth-wise convolutions** has the **lowest latency of 2.52 seconds**. Thus, it can be inferred that depth-wise convolutions can help lower inference times by decreasing the number of parameters.

### C. Model Efficiency Techniques

Models with the least number of parameters derived using depth-wise convolutions are subjected to model efficiency techniques mentioned above in the research methodology section. In general, these techniques improve model performance in terms of overall accuracy, with **Blurpool** being the most efficient and consistent of them. Table 4 below shows the improvement seen in the accuracy for various techniques that have been implemented in our work.

Table 4. Effect of various techniques on model accuracy

| Methodologies using Depth-wise convolutions and GAP | Validation Accuracy | |
|---|---|---|
| | MNIST | CIFAR10 |
| | 1,560 Params | 143,396 params |
| Baseline Model | 98.35 | 79.9 |
| One Cycle LR | 98.92 | 79.42 |
| Cutout (CO) | 98.45 | 82.92 |
| Blurpool (BP) | **99.21** | 83.16 |
| Squeeze and excite (SE) | 99.18 | 81.61 |
| Mixup (M) | 97.85 | **83.52** |
| Label Smoothing (LS) | 98.83 | 81.06 |
| Sharpness aware minimization (SAM) | 98.77 | 80.64 |
| Stochastic weight averaging (SWA) | 99.02 | 82.1 |
| BP + SE + SWA | **99.2** | - |

Table 5. Combination of Techniques on Accuracy - CIFAR 10

| Combination techniques using Depth-wise convolutions and GAP – CIFAR10 | Validation Accuracy |
|---|---|
| BP + CO + M + SE | 85.55 |
| BP + CO + M | **86.08** |
| BP + CO + M + SAM | 86.09 |
| BP + CO + M + SWA | 86.37 |
| BP + CO + M + LS | 86.37 |
| BP + CO + M + LS + SWA + SAM | **86.76** |

Performance on MNIST

**Blurpool is the most efficient** model enhancement technique in the MNIST dataset, where there was an increase in accuracy from 98.35% (baseline) to **99.21%** with just **1.5K** training parameters. Furthermore, the top three performing techniques (Blurpool, Squeeze and excite, Stochastic Weight Averaging) were further combined to study the performance improvement; however, the accuracy remained static at 99.2% (Table 4). Hence no further combination techniques were experimented upon.

Performance on CIFAR-10

Similar performance improvements were noted on the CIFAR-10 dataset with techniques such as Mixup, Blurpool, and Cutouts. **Mixup fared better** with a **3.62 % increase** in accuracy from 79.9% to 83.52%.

Inspired by this leap in accuracy with just a single technique, we further combined the top-performing techniques in varying combinations to push the model to attain higher accuracies. It was seen that (Table 5), combining BP + CO + M + LS + SWA + SAM, resulted in the model reaching a **SOTA accuracy of 86.76%** (**6.86% increase**) with just over 140K parameters.

### D. Training Curves

The training curves for various depth-wise convolution models are shown below.

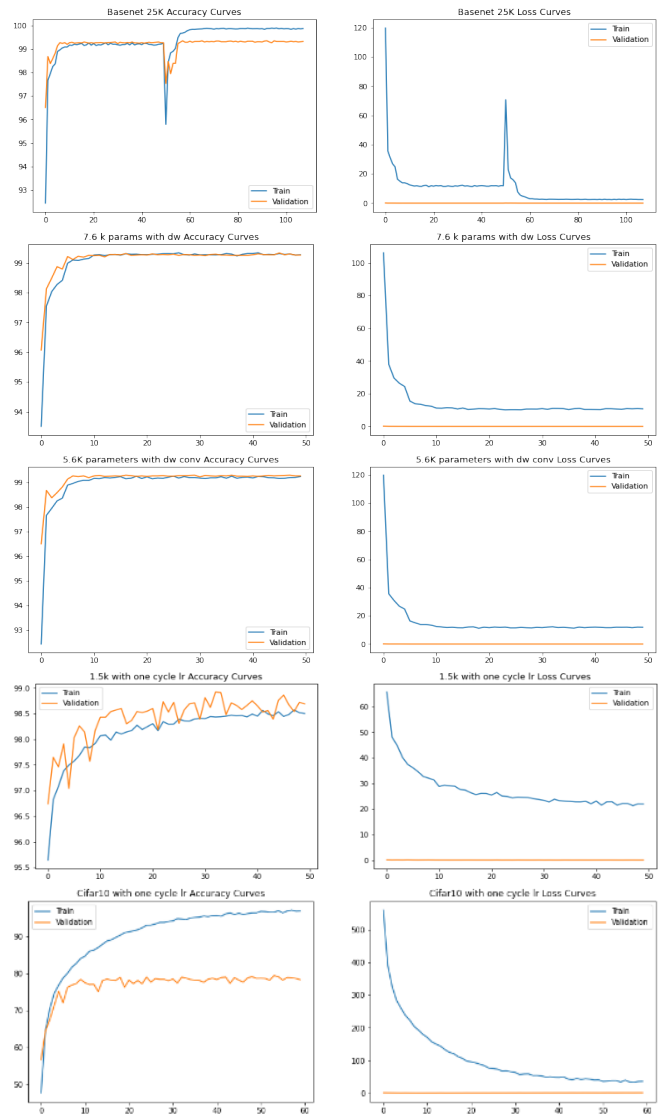

Fig 6. 25K, 7K,5K,1.5K (MNIST),143K model (CIFAR-10) (Top to Bottom) - Training curves

## V. CONCLUSION

Our work sheds light on the efficiency of Depth wise separable convolutions and GAP. These techniques result in a significant reduction in the number of parameters. Though they cannot match the feature extraction capabilities of a conventional 3x3 kernel, fine-tuning the architecture with these techniques will result in equivocal if not better performances of these models, making them ideal for deployment on Edge and mobile devices. As seen in our study, we achieved a SOTA accuracy of 98.35% on the MNIST dataset using just over 1.5K parameters and a model size of 19 KB.

Furthermore, it was observed that combining various model enhancement techniques resulted in better accuracies, as in the case of MNIST, where we achieved an accuracy of 98.35 % using Blurpool. On CIFAR10, the performance gains are even more substantial, with Cutout, Mixup, Blurpool, Label Smoothing, Stochastic Weight Averaging, and Squeeze & Excitation block applied in isolation. Combining all the six techniques, resulted in a phenomenal 6.86% increase from 77.9% to 86.76%. Though Depth wise convolutions do not directly affect the inference time of a model, their ability to reduce the number of parameters helps decrease the overall mathematical computational time required at the time of inference, hence speeding up the model.

## REFERENCES

1. Alex Krizhevsky, Ilya Sutskever, and Geoffrey E Hinton. 2012. Imagenet classification with deep convolutional neural networks. Advances in neural information processing systems 25 (2012), 1097–1105
2. Karen Simonyan and Andrew Zisserman. 2014. Very deep convolutional networks for large-scale image recognition. arXiv preprint arXiv:1409.1556 (2014).
3. Christian Szegedy, Wei Liu, Yangqing Jia, Pierre Sermanet, Scott Reed, Dragomir Anguelov, Dumitru Erhan, Vincent Vanhoucke, and Andrew Rabinovich. 2015. Going deeper with convolutions. In Proceedings of the IEEE conference on computer vision and pattern recognition. 1–9
4. Kaiming He, Xiangyu Zhang, Shaoqing Ren, and Jian Sun. 2016. Deep residual learning for image recognition. In Proceedings of the IEEE conference on computer vision and pattern recognition. 770–778.
5. F. Chollet. Xception: Deep learning with depthwise separable convolutions. arXiv preprint arXiv:1610.02357, 2016.
6. A. G. Howard, M. Zhu, B. Chen, D. Kalenichenko, W. Wang,T. Weyand, M. Andreetto, and H. Adam. Mobilenets: Efficient convolutional neural networks for mobile vision applications. arXiv preprint arXiv:1704.04861, 2017
7. Lin, M.; Chen, Q.; and Yan, S. 2013. Network in network. arXiv preprint arXiv:1312.4400.
8. Zhou, B.; Khosla, A.; Lapedriza, A.; Oliva, A.; and Torralba, A.2016. Learning deep features for discriminative localization. In Computer Vision and Pattern Recognition (CVPR), 2016 IEEE Conference on, 2921–2929. IEEE.
9. Richard Zhang. Making convolutional networks shift-invariant again. In ICML, 2019
10. J. Hu, L. Shen, and G. Sun. Squeeze-and-excitation networks. arXiv preprint arXiv:1709.01507, 2017.
11. Ethiraj S, Bolla BK. Classification Of Astronomical Bodies By Efficient Layer Fine-Tuning Of Deep Neural Networks n.d. https://doi.org/10.1109/CICT53865.2020.9672430.
12. Dileep P, Bolla BK, Ethiraj S. Revisiting Facial Key Point Detection: An Efficient Approach Using Deep Neural Networks 2022. https://doi.org/10.48550/ARXIV.2205.07121
13. Bolla BK, Kingam M, Ethiraj S. Efficient Deep Learning Methods for Identification of Defective Casting Products 2022. https://doi.org/10.48550/ARXIV.2205.07118.
14. T. DeVries and G. W. Taylor. Improved regularization of convolutional neural networks with cutout. arXiv preprint arXiv:1708.04552, 2017

15. H. Zhang, M. Cisse, Y. N. Dauphin, and D. Lopez-Paz. mixup: Beyond empirical risk minimization. arXiv preprint arXiv:1710.09412, 2017.
16. Pavel Izmailov, Dmitrii Podoprikhin, Timur Garipov, Dmitry Vetrov, and Andrew Gordon Wilson. Averaging weights leads to wider optima and better generalization. arXiv preprint arXiv:1803.05407, 2018.
17. Pierre Foret, Ariel Kleiner, Hossein Mobahi, and Behnam Neyshabur. Sharpness-aware minimization for efficiently improving generalization. Arxiv, 2010.01412, 2020
18. C. Szegedy, V. Vanhoucke, S. Ioffe, J. Shlens, and Z. Wojna. Rethinking the inception architecture for computer vision. In CVPR, 2016
19. Team TMM. composer [Internet]. 2021. Available from: https://github.com/mosaicml/composer/